\documentclass[11pt,a4paper]{article}
\usepackage[hyperref]{acl2020}
\usepackage{times}
\usepackage{latexsym}

\usepackage{graphicx}
\usepackage{times}
\usepackage{mathtools}
\usepackage{latexsym}
\usepackage{amsmath}
\usepackage{multirow}
\usepackage{textcomp}
\usepackage{url}
\usepackage{adjustbox}
\usepackage{microtype}
\usepackage{tabularx}
\usepackage{graphicx}
\usepackage{amsmath}
\usepackage{dirtytalk}
\usepackage{enumitem}
\usepackage{booktabs}

\usepackage{microtype}

\aclfinalcopy 

\title{  Multichannel LSTM-CNN for Telugu Technical Domain Identification}

\author{Sunil Gundapu \\
  Language Technologies Research Centre \\
  KCIS, IIIT Hyderabad \\
  Telangana, India \\
  {\tt sunil.g@research.iiit.ac.in} \\\And
  Radhika Mamidi \\
  Language Technologies Research Centre \\
  KCIS, IIIT Hyderabad \\
  Telangana, India \\
  {\tt radhika.mamidi@iiit.ac.in} \\}

\date{}

\begin{document}
\maketitle
\begin{abstract}
With the instantaneous growth of text information, retrieving domain-oriented information from the text data has a broad range of applications in Information Retrieval and Natural language Processing. Thematic keywords give a compressed representation of the text. Usually, Domain Identification plays a significant role in Machine Translation, Text Summarization, Question Answering, Information Extraction, and Sentiment Analysis. In this paper, we proposed the Multichannel LSTM-CNN methodology for Technical Domain Identification for Telugu. This architecture was used and evaluated in the context of the ICON shared task ``TechDOfication 2020" (task h), and our system got 69.9\% of the F1 score on the test dataset and 90.01\% on the validation set.
\end{abstract}

\section{Introduction}

Technical Domain Identification is the task of automatically identifying and categorizing a set of unlabeled text passages/documents to their corresponding domain categories from a predefined domain category set. The domain category set consists of 6 category labels: Bio-Chemistry, Communication Technology, Computer Science, Management, Physics, and Other. These domains can be viewed as a set of text passages, and test text data can be treated as a query to the system. Domain identification has many applications like Machine Translation, Summarization, Question Answering, etc. This task would be the first step for most downstream applications (i.e., Machine Translation). It decides the domain for text data, and afterward, Machine Translation can choose its resources as per the identified domain.

Majority of the research work in the area of text classification and domain identification has been done in English. There has been well below contribution for regional languages, especially Indian Languages. Telugu is one of India's old traditional languages, and it is categorized as one of the Dravidian language family. According to the Ethnologue\footnote{https://www.ethnologue.com/guides/ethnologue200}  list, there are about 93 million native Telugu speakers, and it ranks sixteenth most-spoken languages worldwide.

We tried to identify the domain of Telugu text data using various Supervised Machine Learning and Deep Learning techniques in our work. Our Multichannel LSTM-CNN method outperforms the other methods on the provided dataset. This approach incorporates the advantages of CNN and Self-Attention based BiLSTM into one model.

The rest of the paper is structured as follows: Section 2 explains some related works of domain identification, Section 3 describes the dataset provided in the shared task, Section 4 addresses the methodology applied in the task, Section 5 presents the results and error analysis, and finally, Section 6 concludes the paper as well as possible future works.

\section{Related Work}

Several methods for domain identification and text categorization have been done on Indian languages, and few of the works have been reported on the Telugu language. In this section, we survey some of the methodologies and approaches used to address domain identification and text categorization.

\citet{Kavi:95} explains the automatic text categorization with special emphasis on Telugu. In his research work, supervised classification using the Naive Bayes classifier has been applied to 800 Telugu news articles for text categorization. \citet{Swamy:14} work on representing and categorizing Indian language text documents using text mining techniques K Nearest Neighbour, Naive Bayes, and decision tree classifier. 

Categorization of Telugu text documents using language-dependent and independent models proposed by \citet{Narala:17}. \citet{Durga:11} introduced a model for document classification and text categorization. In their paper described a term frequency ontology-based text categorization for Telugu documents. Combining LSTM and CNN's robustness, \citet{Zhenyu:20} proposed Attention-based Multichannel Convolutional Neural Network for text classification. In their network, BiLSTM encodes the history and future information of words, and CNN capture relations between words.

\section{Dataset Description}
We used the dataset provided by the organizers of Task-h of TechDOfication 2020 for training the models. The data for the task consists of 68865 text documents for training, 5920 for validation, 2611 for testing. For hyperparameter tuning, we used the validation set provided by the organizers. The statistics of the dataset are shown in Table 1. And the amount of texts for the dataset can be seen in Figure 1.

\begin{table}[h!]
  \begin{center}
    \begin{tabular}{c|c|c}
      \hline
      \textbf{Labels($\downarrow$)} & \textbf{Train Data} & \textbf{Validation Data} \\
      \hline
      cse & 24937 & 2175 \\ 
      phy & 16839 & 1650 \\ 
      com\_tech & 11626 & 970 \\ 
      bio\_tech & 7468 & 580 \\ 
      mgnt & 2347 & 155 \\ 
      other & 5648 & 390 \\ 
      \hline
      Total & 68865 & 5920 \\
      \hline
    \end{tabular}
  \end{center}
  \caption{\label{font-table} Dataset Statistics }
\end{table}

\begin{figure}[h!]
  \includegraphics[width=\linewidth]{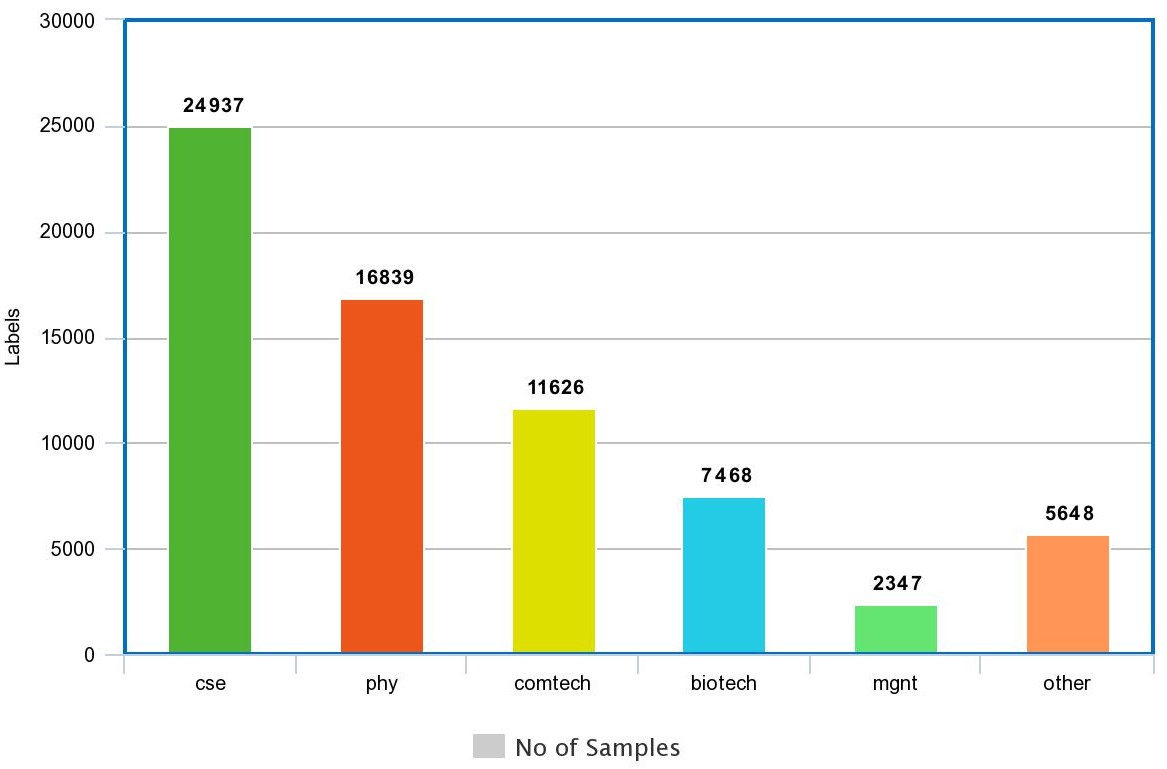}
  \caption{Number of samples per class}
  \label{fig:boat1}
\end{figure}

\section{Proposed Approach}
\subsection{Data Preprocessing}

The text passages have been originally provided in the Telugu script with the corresponding domain tags. The text documents have some noise, so before passing the text to the training stage, they are preprocessed using the following procedure:

\begin{itemize}
\item \textbf{Acronym Mapping Dictionary:} We created an acronym mapping dictionary. Expanded the English acronyms using the acronym mapping dictionary.

\item \textbf{Find Language Words:} Sometimes, English words are co-located with Telugu words in the passage. We find the index of those words to translate into Telugu. 

\item \textbf{Translate English Words:} Translate the English words into the Telugu language, which are identified in the first stage of preprocessing. Google’s Translation API\footnote{https://py-googletrans.readthedocs.io/en/latest/} was used for this purpose.

\item \textbf{Hindi Sentence Translation:} We can observe a few Hindi sentences in the dataset. We translated those sentences into Telugu using Google translation tool.

\item \textbf{Noise Removal:} Removed the unnecessary tokens, punctuation marks, non-UTF format tokens, and single length English tokens from the text data.
\end{itemize}

\subsection{Supervise Machine Learning Algorithms}

To build the finest system for domain identification, we started with supervised machine learning techniques then moved to deep learning models. SVM, Multilayer Perceptron, Linear Classifier, Gradient Boosting methods performed very well on the given training dataset. These supervised models trained on the word level, n-gram level, and character level TF-IDF vector representations.

\subsection{Multichannel LSTM-CNN Architecture}

We started experiments with individual LSTM, GRU, CNN models with different word embeddings like word2vec, glove and fasstext. However, ensembling of CNN with self-attention LSTM model gave better results than individual models.

We develop a multichannel model for domain identification consisting of two main components. The first component is a long short time memory \citep{lstm:97} (henceforth, LSTM).  The advantage of LSTM can handle the long term dependencies but does not store the global semantics of foregoing information in a variable-sized vector. The second component is a convolutional neural network \citep{LeCun:89} (henceforth, CNN). The advantage of CNN can capture the n-gram features of text by using convolution filters, but it restricts the performance due to convolutional filters size. By considering the strengths of these two components, we ensemble the LSTM and CNN model for domain identification.

\begin{figure}[h!]
  \includegraphics[width=\linewidth]{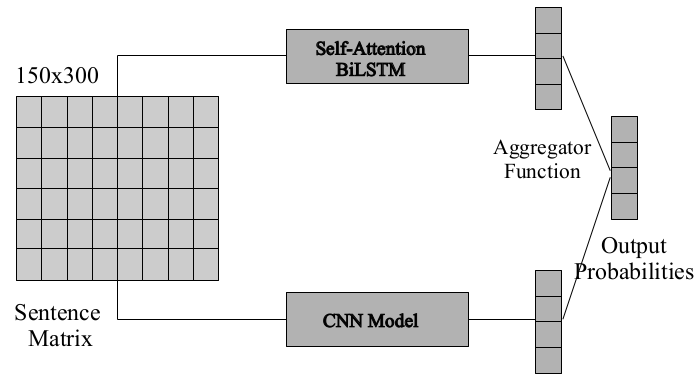}
  \caption{Multichannel LSTM-CNN Model}
  \label{fig:boat1}
\end{figure}

\subsubsection{Self-Attention BiLSTM Classifier}

The first module in architecture is Self-Attention based BiLSTM classifier. We employed this self-attention
\citep{XU:15} based BiLSTM model to extract the semantic and sentiment information from the input text data. Self-attention is an intra-attention mechanism in which a softmax function gives each subword's weights in the sentence. The outcome of this module is a weighted sum of hidden representations at each subword.

\begin{figure}[h!]
  \includegraphics[width=\linewidth]{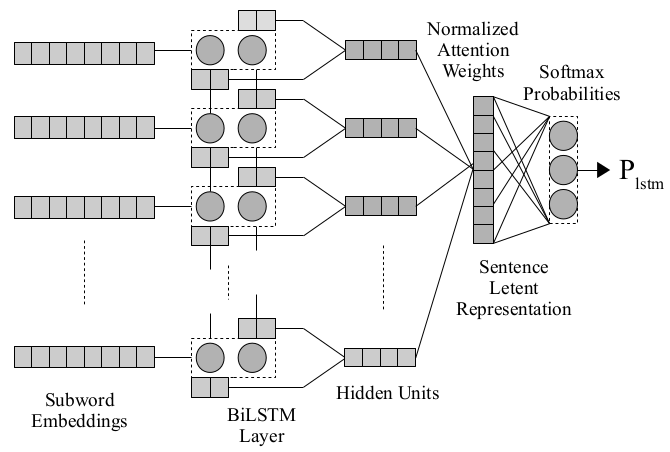}
  \caption{Self-Attention BiLSTM}
  \label{fig:boat1}
\end{figure}

The self-attention mechanism is built on BiLSTMs architecture (See figure 3), and it takes input as pre-trained embeddings of the subwords. We passed the Telugu fasttext \citep{grave:18} subword embeddings to a BiLSTM layer to get hidden representation at each timestep, which is the input to the self-attention component.

Suppose the input sentence $S$ is given by the subwords $(w_1, w_2,...,w_n)$. Let $\overrightarrow{h}$ represents the forward hidden state and $\overleftarrow{h}$ represents the backward hidden state at $i^{th}$ position in BiLSTM. The merged representation $k_i$ is obtained by combining the forward and backward hidden states. We concatenate the forward and backward hidden units to get the merged representations $(k_1, k_2,...k_n)$.

\begin{equation} \label{eq1}
\begin{split}
k_i = [\overrightarrow{h_i};\overleftarrow{h_i}]
\end{split}
\end{equation}

The self-attention model gives a score  $e_i$ to each subword $i$ in the sentence $S$, as given by below equation

\begin{equation} \label{eq1}
\begin{split}
e_i = k_i^Tk_n
\end{split}
\end{equation}

Then we calculate the attention weight $a_i$ by normalizing the attention score $e_i$

\begin{equation} \label{eq1}
\begin{split}
a_i & = \frac{exp(e_i)}{\sum_{j=1}^{n} exp(e_j)}
\end{split}
\end{equation}

Finally, we compute the sentence $S$ latent representation vector $h$ using below equation

\begin{equation} \label{eq1}
\begin{split}
h & = \sum_{i=1}^{a_i \times k_i} 
\end{split}
\end{equation}

The latent representation vector $h$ is fed to a fully connected layer followed by a softmax layer to obtain probabilities $P_{lstm}$.

\begin{figure*}[h!]
 \begin{center}
  \includegraphics[width=11cm, height=6.5cm]{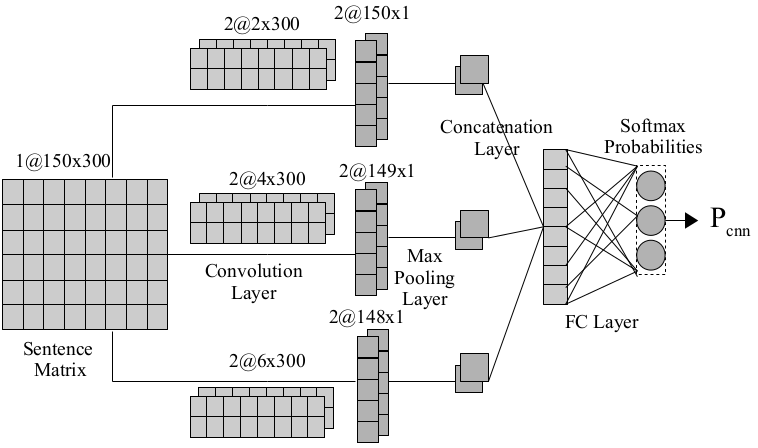}
  \caption{CNN Classifier}
  \label{fig:boat1}
 \end{center}
\end{figure*}

\subsubsection{Convolutional Neural Network}

The second component is CNN, which consider the ordering of the words and the context in which each word appears in the sentence. We present Telugu fasstext subword embeddings (\citet{fasttext:16}) of a sentence to 1D-CNN (See figure 4) to generate the required embeddings.

Initially, we present a $d \times S$ sentence embedding matrix to the convolution layer. Each row is a $d$-dimension fasstext subword embedding vector of each word, and $S$ is sentence length. We perform convolution operations in the convolution layer with three different kernel sizes (2, 4, and 6). The purpose behind using various kernel sizes was to capture contexts of varying lengths and to extract local features around each word window. The output of convolution layers was passed to corresponding max-pooling layers. The max-pooling layer is used to preserve the word order and bring out the important features from the feature map. We change the original max-pooling layer in the convolution neural network with the word order-preserving k-max-pooling layer to preserve the inputted sentences word order. The order persevering max-pooling layer reduces the number of features while preserving the order of these words.

The max-pooling layer output is concatenated together fed to a fully connected layer followed by a softmax layer to obtain softmax probabilities $P_{cnn}$.

\begin{table*}[h!]
\centering
\begin{adjustbox}{width=\textwidth}
\begin{tabular}{c cccc cccc}
\toprule
 & \multicolumn{4}{c}{\textbf{Validation Data}} & \multicolumn{4}{c}{\textbf{Test Data}} \\
\cmidrule(lr){2-5} \cmidrule(lr){6-9}
Model    & Accuracy & Precision   & Recall  & F1-Score & Accracy & Precision & Recall & F1-Score \\
\midrule
SVM & 0.8612 & 0.8598 & 0.8601  & 0.866  & -  & - & -  & - \\
CNN & 0.8890 & 0.8874 & 0.8897  & 0.8902  & 0.6465  & 0.6558 & 0.6948  & 0.6609 \\
BiLSTM & 0.8706 & 0.8714 & 0.8702  & 0.8786  & 0.6465  & 0.6519 & 0.6927  & 0.6571 \\
Multichannel & 0.9001 & 0.9008 & 0.9006  & 0.9003 & \textbf{0.6928} & \textbf{0.6984} & 0.7228 & 0.6992 \\
Organizers System & - & - & -  & -  & 0.6855  & 0.6974 & 0.7289  & 0.7028 \\
\bottomrule
\end{tabular}
\end{adjustbox}
\caption{\label{font-table} Comparison between various model results }
\end{table*}

The CNN and BiLSTM models softmax probabilities are aggregated (See figure 2) using an element-wise product to obtain the final probabilities $P_{final} = P_{cnn} \circ P_{lstm}$. We tried various aggregation techniques like average, maximum, minimum,  element-wise addition, and element-wise multiplication to combine LSTM and CNN models' probabilities. But an element-wise product gave better results than other techniques.

\section{Results and Error Analysis}

We first started our experiments with machine learning algorithms with various kinds of TF-IDF feature vector representations. SVM, MLP gave pretty good results on the validation dataset but failed on bio-chemistry and management data points. And CNN model with fasstext word embeddings seemed to be confused between the computer technology and cse data points (See table 2). Maybe the reason behind this confusion is that both datapoints quite similar at the syntactic level.

The self-attention based BiLSTM model outperforms the CNN model on physics, cse, and computer technology data points though it performs worse on the bio-chemistry and management data points. If we observe the training set, 75\% of the data samples belong to physics, cse, and computer technology domains remaining 25\% of data owned by the remaining domain labels. So we were assuming that an imbalanced training set was also one of the problems for misclassification of bio-chemistry and management domain samples. We tried to handle the this data-skewing problem with SMOTE \citep{Chawla:02} technique, but it is doesn't work very well.

\begin{table}[h!]
  \begin{center}
    \begin{tabular}{c|cc}
      \hline
      \textbf{Measures($\downarrow$)} & \textbf{Validation Data} & \textbf{Test Data} \\
      \hline
      precision & 0.90 & 0.7228 \\ 
      recall & 0.90 & 0.6984 \\ 
      f1\-score & 0.90 & 0.6992 \\ 
      accuracy & 0.9003 & 0.6928 \\ 
      \hline
    \end{tabular}
  \end{center}
  \caption{\label{font-table} Performance of multichannel system}
\end{table}

After observing the CNN and BiLSTM model results, we ensemble these two models for better results. The ensemble Multichannel LSTM CNN model outperforms all our previous models, achieving a recall of 0.90 with the weighted F1-score of 0.90 on the development dataset and a recall of 0.698 with an F1-score of 0.699 on the test dataset (See table 3).

\section{Conclusion}

In this paper, we proposed a multichannel approach that integrates the advantages of CNN and LSTM. This model captures local, global dependencies, and sentiment in a sentence. Our approach gives better results than individual CNN, LSTM, and supervised machine learning algorithms on the Telugu TechDOfication dataset.

As discussed in the previous section, we will handle the data imbalance problem efficiently in future work. And we will improve the performance of the unambiguous cases in cse and computer technology domains.

\bibliography{anthology,acl2020}
\bibliographystyle{acl_natbib}

\end{document}